\documentclass[runningheads]{llncs}

\usepackage{eccv}

\usepackage{eccvabbrv}

\usepackage{graphicx}
\usepackage{booktabs}
\usepackage{colortbl} 
\usepackage{xcolor} 

\usepackage[accsupp]{axessibility}  

\usepackage{hyperref}

\usepackage{orcidlink}

\usepackage{float}
\usepackage{placeins}
\begin{document}

\title{OSSA: Unsupervised One-Shot Style Adaptation}


\author{Robin Gerster\inst{1} \orcidlink{https://orcid.org/0000-0002-6210-5735} \and
Holger Caesar\inst{1} \orcidlink{https://orcid.org/0000-0001-5099-6297}  \and
Matthias Rapp\inst{2} \and
Alexander Wolpert\inst{2} \and
Michael Teutsch\inst{2} \orcidlink{https://orcid.org/0000-0001-8739-2702} 
}

\authorrunning{R. Gerster et al.}

\institute{Delft University of Technology, The Netherlands
\and
Hensoldt Optronics GmbH, Germany\\
}

\maketitle

\begin{abstract}
Despite their success in various vision tasks, deep neural network architectures often underperform in out-of-distribution scenarios due to the difference between training and target domain style. To address this limitation, we introduce One-Shot Style Adaptation (OSSA), a novel unsupervised domain adaptation method for object detection that utilizes a single, unlabeled target image to approximate the target domain style. Specifically, OSSA generates diverse target styles by perturbing the style statistics derived from a single target image and then applies these styles to a labeled source dataset at the feature level using Adaptive Instance Normalization (AdaIN). Extensive experiments show that OSSA establishes a new state-of-the-art among one-shot domain adaptation methods by a significant margin, and in some cases, even outperforms strong baselines that use thousands of unlabeled target images. By applying OSSA in various scenarios, including weather, simulated-to-real (sim2real), and visual-to-thermal adaptations, our study explores the overarching significance of the style gap in these contexts. OSSA's simplicity and efficiency allow easy integration into existing frameworks, providing a potentially viable solution for practical applications with limited data availability. Code is available at \url{https://github.com/RobinGerster7/OSSA}

\end{abstract}

\section{Introduction}
\label{sec:intro}
Effective object detection across various deep neural network architectures hinges on generating a compact yet rich feature representation \cite{9710354}. Traditionally, this task has relied on Convolutional Neural Networks (CNNs) as image feature extractors. However, the effectiveness of CNNs is critically dependent on the assumption that the training and test datasets come from the same distribution \cite{Hendrycks, Recht}. This assumption is rarely met in practical applications, which are influenced by a variety of factors, including changing weather conditions, shifts from simulated to real environments, and cross-spectral domain variations. This fundamental discrepancy between the source domain, on which the detector was trained, and the target domain, where it is deployed, can lead to significant performance degradation — an issue of great concern in high-stakes fields such as autonomous driving \cite{autonomous} or medical diagnostics \cite{medicalrisks}.

Research on standard CNNs has revealed that their inductive bias differs significantly from human vision. While humans tend to recognize objects based on their shape, CNNs exhibit a strong bias towards styles and textures~\cite{Geirhos, texturebiasorigins}. Given that image styles are sensitive to changes across domains, this explains why CNNs are prone to suffering from domain shifts. Geirhos et al.~\cite{Geirhos} supported this hypothesis by demonstrating that CNNs trained with heavy style augmentation become more robust against various image distortions. Similarly, Nam et al.~\cite{reducingstylebias} showed that reducing the style bias in CNNs can enhance their robustness in cross-domain detection.

\begin{figure}[t] 
\centering
\includegraphics[width=\textwidth]{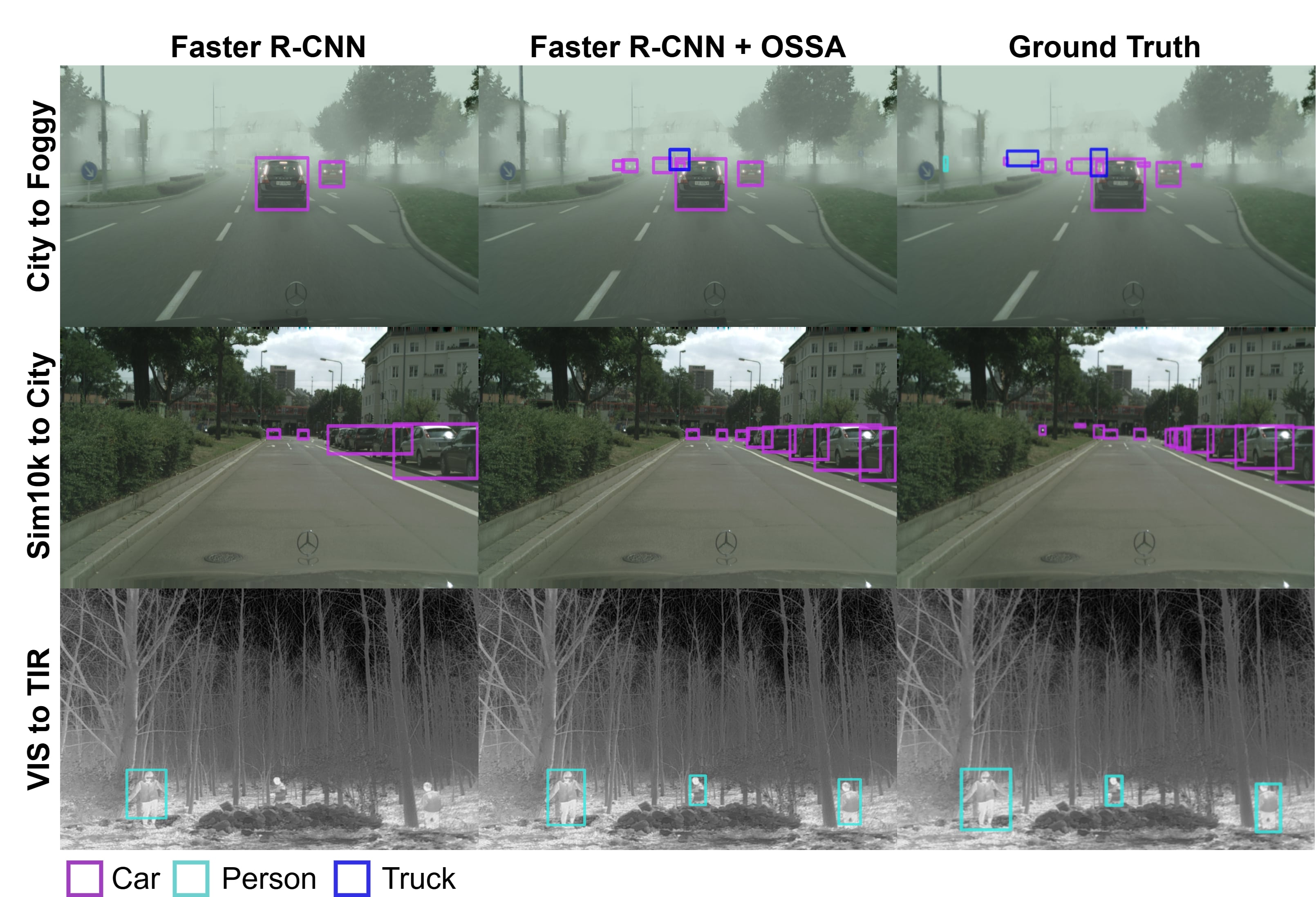} 
\caption{A qualitative comparison of the baseline Faster R-CNN, our proposed OSSA, and the ground truth. We observe that in various scenarios, including weather (top row), sim2real (middle row), and visual-optical to thermal infrared adaptation, OSSA leads to a substantial increase in accurate detections.}
\label{fig:qual}
\end{figure}

 To overcome the domain shift associated with the style gap current research mostly presupposes the availability of extensive target domain datasets~\cite{ViSGA, IRGG, GIPA, SC-UDA}, or in-depth prior knowledge of the target domain~\cite{CLIP}. State-of-the-art efforts to bridge the style gap often involve using complex generative models for image-level style augmentation~\cite{ProCST, TLDR}, which are usually inefficient and slow, particularly with larger images. Conversely, adapting styles at the feature map level has proven to be more effective in overcoming the style gap and does not significantly impact training time~\cite{NP, DSU, Mixstyle}.

The goal of adaptation strategies is to move towards techniques that bypass the need for large-scale data collection, which can be costly, time-consuming, or otherwise impractical in many settings. One-shot adaptation represents an extreme case of data frugality and, thus, an advancement towards more agile and adaptable object detection in practical applications. The need for frugal learning can be observed in various fields including but not limited to sim2real adaptation for underwater image enhancement~\cite{underwater}, medical imaging~\cite{ZHANG2024102996}, and hyperspectral image segmentation~\cite{spectral}.

Our contributions can be summarized as follows: (1)~We propose a simple, unsupervised one-shot style adaptation technique, establishing a new state-of-the-art for object detection in this context. (2)~We show that a single image can effectively capture the style of a target dataset, significantly reducing the need for extensive data collection. (3)~We explore the importance of addressing the style gap in diverse contexts including weather, synthetic-to-real and visual-optical to thermal infrared adaptation.

\section{Related Work}
\label{sec:background}

\subsection{Domain Gaps in Object Detection}
The performance of object detectors often declines due to distribution shifts between datasets, a problem commonly referred to as the domain gap. Recent works have begun to disentangle this domain gap into appearance and content gaps \cite{metasim, CARE}. The appearance gap consists of visual disparities between images from the two domains, such as differences in lighting between real and simulated images at the pixel level \cite{pasta}, or variations in the appearance of synthesized versus real objects at the instance level. The content gap, on the other hand, refers to disparities in task label distributions and scene layouts. Additionally, the style gap \cite{SC-UDA, jiang, stylecons, Li_2023_WACV}, while synonymous to the appearance gap, is understood to not significantly contribute semantic information to the object detection task \cite{Li_2022_CVPR}. Consequently, many studies aim to establish style invariances under the assumption that style does not contribute essential information to the object recognition and detection tasks \cite{NP, DSU, Mixstyle, pAdaIN}.

\subsection{Unsupervised Domain Adaptation}
Unsupervised Domain Adaptation (UDA) in transfer learning aims to learn knowledge that generalizes across domains with varying distributions. Typically, UDA assumes access to a labeled source domain and an unlabeled target domain that follows a different distribution. The primary challenge in UDA is mitigating the domain shift, or the discrepancy between the marginal probability distributions of these different domains \cite{yang2023tvt,Oza2024}.

Several strategies for UDA have been developed. One popular strategy is domain adversarial learning, which seeks to learn domain-invariant representations by employing a discriminator trained to differentiate between domains based on input features \cite{yang2023tvt}. This adversarial approach is central to many methods such as DA-Faster~\cite{DAFRCNN}, D\&Match~\cite{DMatch}, VisGA~\cite{ViSGA}, and GPA~\cite{GIPA}. Another more recent approach that has achieved good results is domain adaptation using the mean-teacher framework. This framework employs a student-teacher setup, where the student network is trained to replicate the teacher network's outputs, which are smoothed over time by taking the exponential moving average of the student network's parameters, providing a robust means to adapt to new domains without requiring labeled target data. Mean-teacher is the backbone of methods including MTOR \cite{MTOR}, MKT \cite{MKT}, and the current state-of-the-art TDD \cite{TDD}.

However, a common issue with existing methods, including those mentioned above, is their dependence on a substantial amount of unlabeled target data. This reliance is often impractical due to constraints like the cost or logistics associated with data collection \cite{howmuchrealdataisactuallyneeded}. 

\subsection{One-Shot Adaptation}
One-shot adaptation, in the context of UDA, refers to the task of adapting a model using only the available source data and a single unlabeled target image. This approach addresses the challenge of minimal target data, making it a valuable area of study. However, in the field of object detection, only a handful of such strategies have been developed. The most well-known methods include One-Shot Feature Alignment (OSFA) \cite{OSFA}, One-Shot Adaptive Cross-Domain Detection (OSHOT) \cite{OSHOT}, and Self-Supervision and Meta-Learning for One-Shot Unsupervised Cross-Domain Detection (FULL-OSHOT) \cite{FULL-OSHOT}.

The OSHOT strategy involves two phases aimed at improving object detection. First, during pretraining, a standard detection model is enhanced with an image rotation classifier. This auxiliary task helps the model learn adaptable features that generalize better across different object orientations. Second, in the adaptation stage, network features are refined using a single target sample, focusing on optimizing the rotation objective. Additionally, OSHOT employs pseudo-labeling to enhance the auxiliary task's focus on local object context.
The FULL-OSHOT approach extends OSHOT with a second pretraining stage involving meta-learning with rotation recognition as the inner optimization task. This prepares the network for test-time adaptation, where both rotation and feature extractor modules are iteratively updated on a single test sample.
The OSFA method proposes a one-shot feature alignment algorithm. Specifically, OSFA reduces the domain gap by introducing a novel domain loss. This loss penalizes differences in the mean activation of the first VGG16 pooling layer between a source, training image and a single target image.

\section{One-Shot Style Adaptation}
\label{sec:ossa}

The consensus in current research is that CNNs not only encode style information~\cite{Gatys}, but do so predominantly in the early layers~\cite{Mixstyle, TF-CAL, pan}. Based on this insight, various domain generalization methods have been developed, utilizing channel style statistics and Adaptive Instance Normalization (AdaIN) at these initial layers to enhance generalization across different domains~\cite{NP, Mixstyle, pAdaIN, DSU}.

AdaIN is a technique used to transfer the style from one image onto another by adjusting the statistics of the source feature maps to match those of the target style feature map. Specifically, AdaIN takes a source feature map, representing the content to be stylized, and a target style feature map, representing the style to be transferred. AdaIN can also facilitate domain adaptation by transferring the style of an unlabeled target domain of interest onto labeled source domain images for which labels are available. The process involves normalizing the source feature maps $x$ so that its mean and variance match those of target style $y$. Following Huang and Belongie~\cite{AdaIN}, this operation can be defined as:

\begin{equation}
\text{AdaIN}(x, y) = \sigma(y) \left( \frac{x - \mu(x)}{\sigma(x)} \right) + \mu(y)
\end{equation}

Here, \( \mu(x) \) and \( \sigma(x) \) refer to the mean and standard deviation of the input feature maps \( x \), while \( \mu(y) \) and \( \sigma(y) \) refer to those of the target style image \( y \). Statistics are calculated across the spatial dimensions (height \( H \) and width \( W \)) for each channel \( C \) in each instance \( B \) of the batch.

\begin{equation}
\mu(x)_{b,c} = \frac{1}{HW} \sum_{h=1}^{H} \sum_{w=1}^{W} x_{b,c,h,w}
\end{equation}

\begin{equation}
\sigma(x)_{b,c} = \sqrt{\frac{1}{HW} \sum_{h=1}^{H} \sum_{w=1}^{W} \left( x_{b,c,h,w} - \mu(x)_{b,c} \right)^2}
\end{equation}

\begin{figure}[h] 
\centering
\includegraphics[width=0.85\textwidth]{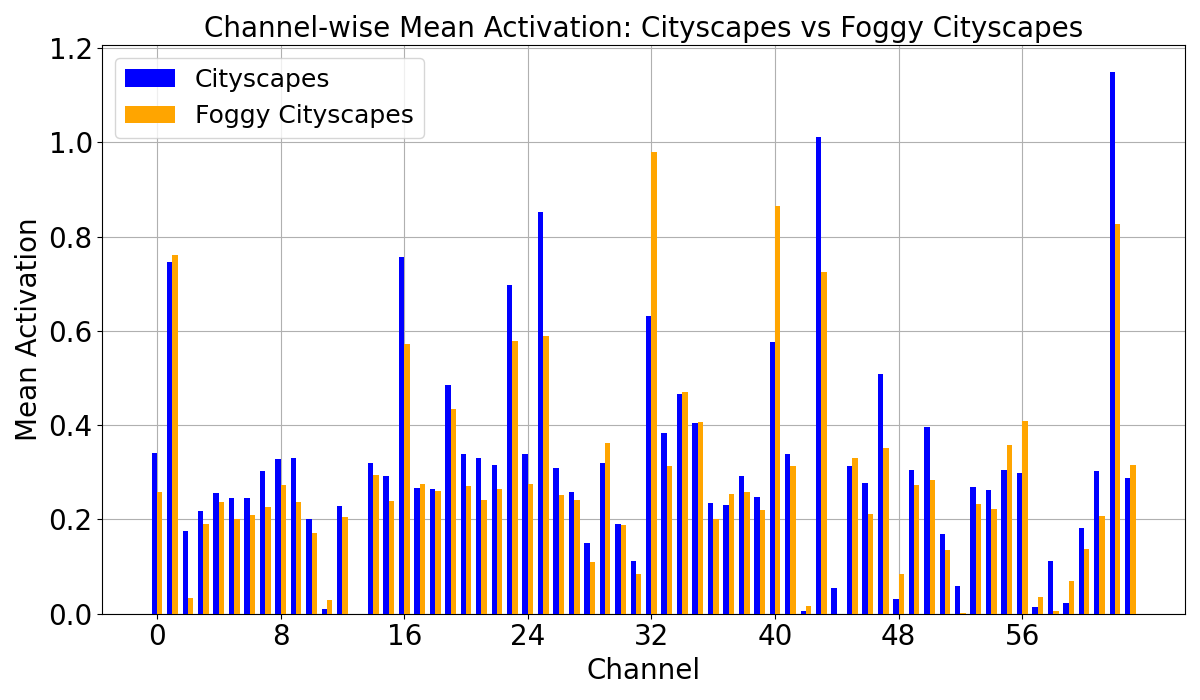} 
\caption{The histogram displays the mean activations for each channel across two datasets, Cityscapes and Foggy Cityscapes. It clearly shows distinct mean activations in each channel at the first layer of ResNet50, indicating a style gap between the datasets.}
\label{fig:histo}
\end{figure}

Our method diverges from prior works by recognizing that approaches, which merely perturb the style of the source dataset to promote domain generalization~\cite{Mixstyle, NP, DSU, pAdaIN, TF-CAL}, are often biased towards the source dataset's style distribution. This bias is illustrated in Fig. \ref{fig:histo}, which demonstrates the significant differences in channel means between datasets like Cityscapes \cite{Cityscapes} and Foggy Cityscapes \cite{Foggy}.

The standard practice of applying multiplicative Gaussian noise \cite{NP, DSU, adversarialstyle}, rather than additive noise, tends to preserve the relative magnitude of the source channel style statistics. The key novelty of OSSA, over prior methods such as NP \cite{NP}, is that OSSA leverages this systematic bias by first gauging the style statistics of the target dataset using a single image. Upon establishing a 'style prototype' of the target, OSSA strategically perturbs around this estimate to generate new, varied target styles.

OSSA then uses AdaIN to transfer the generated target style onto the source image at the feature level. Inspired by the good results of prior works utilizing style perturbation, predominantly NP \cite{NP}, OSSA applies multiplicative Gaussian noise centered at a mean of 1 to the target channel style statistics. This ensures that the generated novel styles remain close to the target. Formally, OSSA can be understood as:

\begin{equation}
\text{OSSA}(x_i, y) = \alpha \sigma(y) \left( \frac{x_i - \mu(x_i)}{\sigma(x_i)} \right) + \beta \mu(y), \quad \alpha, \beta \sim \mathcal{N}(1, 0.75)
\end{equation}

\begin{figure*}[h]
\centering
\includegraphics[width=\textwidth]{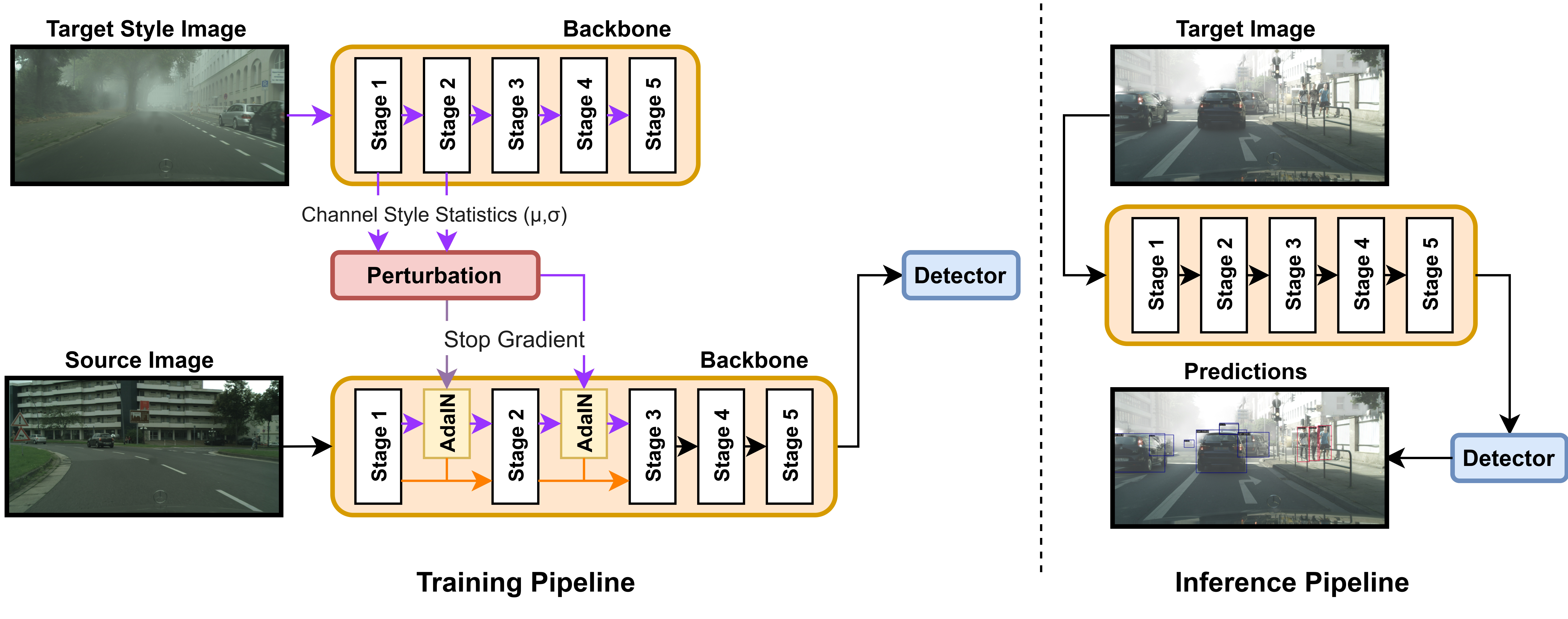}
\caption{High-level overview of the OSSA (One-Shot Style Adaptation) pipeline. When OSSA is active (purple paths), the target image style is extracted from the first two layers of the ResNet50 backbone network, then perturbed with multiplicative Gaussian noise \( \sim \mathcal{N}(1, 0.75) \). The novel style is then integrated at the feature map level of the source images using AdaIN for training. When OSSA is inactive (orange paths), a standard pipeline is followed; black paths indicate components that are always active. The target style statistics $\mu$ and $\sigma$ only need to be computed once.}
\label{fig:ossaarchitecture}
\end{figure*}

In practice, OSSA is applied to the first two layers of the ResNet50 \cite{resnet} (\(i \in \{1,2\}\)), specifically after the pre-convolutional layer. Additionally, OSSA is applied with a probability of 50\%; in other cases, the model undergoes regular training without OSSA. This probabilistic application is crucial as it allows for leveraging the high-quality data of the source domain while simultaneously enhancing the model's performance in the target domain. A visual overview of OSSA is presented in Fig. \ref{fig:ossaarchitecture}.

\section{Experimental Setup}
\label{sec:setup}

In this section, we outline the datasets and implementation details used in our experiments. We utilize several well-known datasets to assess our approach under diverse conditions and provide a thorough description of our implementation methodology to ensure reproducibility.

\subsection{Datasets}

\textbf{Cityscapes \cite{Cityscapes}}: This dataset comprises urban street scenes with pixel-level annotations of 8 object categories. It contains 2,975 images in the training split and 500 images in the validation split. We use the instance level pixel annotations to generate bounding
boxes of objects, as in \cite{DAFRCNN}.

\noindent\textbf{Foggy Cityscapes \cite{Foggy}}: Developed by adding synthetic fog to the original Cityscapes images, Foggy Cityscapes challenges object detection algorithms by simulating adverse weather conditions. For our study, we use the images with the highest level of artificial fog. Here the data split is identical to the Cityscapes dataset where 2,975 images are used for training and 500 images for testing. We use the same procedure as for the Cityscapes dataset to generate bounding box annotations from the semantic masks. 

\noindent\textbf{Sim10k \cite{Sim10k}}: The Sim10k dataset originates from the game Grand Theft Auto V (GTA V) and includes 10,000 synthetic images with corresponding bounding box annotations. This dataset exclusively labels the category \emph{car}. 

\noindent\textbf{M\textsuperscript{3}FD \cite{M3FD}}: Designed for object detection across multi-modal data, the M\textsuperscript{3}FD dataset features 4,200 aligned image pairs from both the visual-optical and thermal infrared spectra, covering diverse scenes. It provides manual annotations for six classes: People, Car, Bus, Motorcycle, and Truck; we exclude the Lamp class from our analysis. The primary reason for this exclusion is that the objects in this class are generally very small, which our Faster R-CNN is not equipped to handle effectively, thereby potentially confounding the analysis. Given that M\textsuperscript{3}FD does not include a predefined train-test split, we adopt the split used in \cite{deevi2024rgb, liang2023explicit}.

\subsection{Implementation Details}
In our implementation, we primarily align with the methodologies established by prior one-shot adaptation methods for object detection \cite{OSHOT, OSFA, FULL-OSHOT}. We utilize the Faster R–CNN architecture \cite{frcnn}  and base our code on the MMDetection framework \cite{mmdetection}. The shortest dimension of the input images is resized to 600 pixels while keeping the aspect ratio fixed. Our model is trained over 70,000 iterations using Stochastic Gradient Descent (SGD) with a momentum of 0.9. We initialize the learning rate at 0.001 and implement a decay by a factor of 10 after the first 50,000 iterations. To avoid confounds and follow comparable work \cite{NP, OSHOT, FULL-OSHOT}, we keep the first two blocks of the ResNet50 backbone frozen and do not apply batch normalization or image-level augmentations such as horizontal flipping.

\section{Experiments}
\label{sec:experiments}
In this section, we conduct a comprehensive evaluation of various methods across multiple datasets and corresponding adaptation tasks. To ensure fair comparisons, all methods in this study use a ResNet50 backbone and Faster R-CNN detector unless stated otherwise.

Our evaluation spans two established benchmarks for domain adaptation within the visual-optical spectrum: Cityscapes to Foggy Cityscapes (representing adverse weather conditions) and Sim10k to Cityscapes (representing sim2real adaptation). The choice to test on adverse weather scenarios is motivated by the presence of a controlled style gap, given that the images are identical (i.e. no content gap) apart from the synthetic fog. Meanwhile the choice to test on Sim10k to Cityscapes is motivated by the growing interest in synthetic data while the presence of both a style and content related gap also offers a complementary and more challenging setting. Finally, we also test OSSA on a visible-to-thermal benchmark. This means that a deep neural network trained on images in the visual-optical spectrum shall be transferred to the thermal infrared spectrum. This scenario is interesting due to the lack of existing research in this area. The perfect alignment between visual-optical and thermal infrared image pairs minimizes the content gap, allowing us to study the significance of style adaptation in this cross-spectral context. For each experiment we report the mean Average Precision (mAP) at a 50\% Intersection over Union (IoU) threshold, along with standard deviations computed from 10 runs to ensure accurate results. Each run involves randomly sampling a new style image from the training split of the target dataset.

\subsection{Adverse Weather Adaptation}

Table \ref{tab:foggy_comparison} shows that our method significantly outperforms the Faster R-CNN baseline and previous one-shot methods such as OSFA \cite{OSFA}, OSHOT \cite{OSHOT}, and FULL-OSHOT \cite{FULL-OSHOT}. Notably, it also exceeds the performance of the state-of-the-art, zero-shot single-domain generalization method NP by nearly three percent mAP. This is significant because previous one-shot methods (OSFA, OSHOT, and FULL-OSHOT) did not perform much better than single-domain zero-shot methods, raising questions about the effectiveness of one-shot approaches. However, our results demonstrate that a single shot can make a substantial difference. When comparing OSSA to methods that perform UDA with access to the full target dataset, we note that we outperform not only established weaker baselines such as DA-Faster and D\&Match but also several stronger baselines such as GPA. However, OSSA does not reach the performance of top-tier methods, indicating that while one-shot adaptation shows promise, there is still room for improvement to match the results of methods with full target data access.

\begin{table}[htbp]
\centering
\resizebox{\textwidth}{!}{%
\begin{tabular}{@{}lccccccccccc@{}}
\toprule
Method  & Target Data & Person & Rider & Car & Truck & Bus & Train & Motorcycle & Bicycle & mAP $\uparrow$ \\
\midrule
\rowcolor{lightgray} Oracle & Full & 44.0\scriptsize$\pm$0.3\normalsize  &  49.4\scriptsize$\pm$0.8\normalsize  & 62.7\scriptsize$\pm$0.3\normalsize & 31.0\scriptsize$\pm$1.5\normalsize & 47.3\scriptsize$\pm$1.7\normalsize & 34.7\scriptsize$\pm$3.8\normalsize & 33.0\scriptsize$\pm$0.8\normalsize & 43.8\scriptsize$\pm$0.8\normalsize & 43.2\scriptsize$\pm$0.5\normalsize \\
\midrule
\multicolumn{10}{c}{Full Target Data Methods} \\
\midrule
DA-Faster \cite{DAFRCNN} & Full & 29.2 & 40.4 & 43.4 & 19.7 & 38.3 & 28.5 & 23.7 & 32.7 & 32.0 \\
D\&Match \cite{DMatch} & Full & 31.8 & 40.5 & 51.0 & 20.9 & 41.8 & 34.3 & 26.6 & 32.4 & 34.9 \\
MTOR \cite{MTOR} & Full & 30.6 & 41.4 & 44.0 & 21.9 & 38.6 & 40.6 & 23.8 & 36.5 & 35.1 \\
SWDA \cite{SWDA} & Full & 31.8 & 44.3 & 48.9 & 21.0 & 43.8 & 28.0 & 28.5 & 32.9 & 35.3 \\
SC-DA \cite{SCDA} & Full & 33.8 & 42.1 & 52.1 & 26.8 & 42.5 & 26.5 & 29.2 & 34.5 & 35.9 \\
IRG \cite{IRGG} & Full & 37.4 & 45.2 & 51.9 & 24.4 & 39.6 & 25.2 & 31.5 & 41.6 & 37.1 \\
GPA \cite{GIPA} & Full & 32.9 & 46.7 & 54.1 & 24.7 & 45.7 & 41.1 & 32.4 & 38.7 & 39.5 \\
DSS \cite{DSS} & Full & 42.9 & 51.2 & 53.6 & 33.6 & 49.2 & 18.9 & 36.2 & 41.8 & 40.9 \\
AFAN \cite{AFAN} & Full & 42.5 & 44.6 & 57.0 & 26.4 & 44.1 & 37.1 & 39.7 & 39.6 & 41.4 \\
ViSGA \cite{ViSGA} & Full & 38.8 & 45.9 & 57.2 & 29.9 & 50.2 & 51.9 & 31.9 & 40.9 & 43.3 \\
MKT \cite{MKT} & Full & 43.5 & 52.0 & 63.2 & 34.7 & 52.7 & 45.8 & 37.1 & 49.4 & 47.3 \\
TDD \cite{TDD} & Full & 50.7 & 53.7 & 68.2 & 35.1 & 53.0 & 45.1 & 38.9 & 49.1 & 49.2 \\
\midrule
\multicolumn{10}{c}{Zero-Shot \& One-Shot Methods} \\
\midrule
Faster R-CNN \cite{frcnn} & Zero & 32.1\scriptsize$\pm$0.5\normalsize & 39.1\scriptsize$\pm$1.1\normalsize & 40.4\scriptsize$\pm$0.7\normalsize & 19.2\scriptsize$\pm$2.3\normalsize & 29.3\scriptsize$\pm$1.9\normalsize & 8.6\scriptsize$\pm$3.7\normalsize & 21.0\scriptsize$\pm$1.3\normalsize & 35.4\scriptsize$\pm$0.7\normalsize & 28.1\scriptsize$\pm$0.9\normalsize \\
NP \cite{NP} & Zero & 41.5 & 48.8 & 55.1 & 22.2 & 39.9 & 15.8 & 28.6 & 44.4 & 37.0 \\
OSFA (VGG16+SSD) \cite{OSFA} & One & 23.6 & 32.6 & 43.8 & 22.9 & 35.4 & 14.7 & 23.1 & 33.2 & 28.7 \\
FULL-OSHOT \cite{FULL-OSHOT} & One & 32.0 & 39.7 & 43.8 & 18.8 & 31.8 & 10.6 & 22.1 & 33.2 & 29.0 \\
OSHOT \cite{OSHOT} & One & 32.1 & 46.1 & 43.1 & 20.4 & 39.8 & 15.9 & 27.1 & 32.4 & 31.9 \\
OSSA (Ours) & One & \textbf{42.0}\scriptsize$\pm$0.7\normalsize & \textbf{49.1}\scriptsize$\pm$0.6\normalsize & \textbf{55.4}\scriptsize$\pm$1.2\normalsize & \textbf{28.4}\scriptsize$\pm$2.1\normalsize & \textbf{42.2}\scriptsize$\pm$3.3\normalsize & \textbf{25.1}\scriptsize$\pm$4.0\normalsize & \textbf{31.1}\scriptsize$\pm$2.2\normalsize & \textbf{44.9}\scriptsize$\pm$0.6\normalsize & \textbf{39.8}\scriptsize$\pm$1.1\normalsize \\
\bottomrule
\end{tabular}%
}
\vspace{0.3cm}
\caption{Comparison of different UDA methods on Cityscapes to Foggy Cityscapes (adverse weather) adaptation. Unless stated otherwise, all methods use a ResNet50+Faster R-CNN architecture.}
\label{tab:foggy_comparison}
\end{table}

\FloatBarrier

\subsection{Sim2Real Adaptation}

In the sim2real adaptation scenario as indicated in Table \ref{tab:sim10k_comparison}, our method attains a high mAP of 52.7, outperforming not only previous one-shot methods and the state-of-the-art single-domain generalization method NP, but also most of the presented full-target methods. However, it is important to note that we do not narrow the gap to the oracle as substantially as we did in the adverse weather scenario. This can be attributed to the the existence of a pronounced content gap \cite{contentgap}. Given that OSSA focuses exclusively on bridging a style gap this leaves the content gap unaddressed. Interestingly, despite only addressing a global style gap, OSSA achieves results competitive with methods that not just utilize the full target dataset but also address multiple types of domain gaps.  This provides evidence that OSSA is an effective tool for unsupervised one-shot style adaptation in the context of sim2real adaptation assuming a style gap exists.

\begin{table}[htbp]
\centering
\resizebox{0.5\textwidth}{!}{%
\begin{tabular}{@{}lcc@{}}
\toprule
Method & Target Data & Car AP $\uparrow$ \\
\midrule
\rowcolor{lightgray} Oracle & Full & 67.7\scriptsize$\pm$0.4\normalsize \\
\midrule
\multicolumn{3}{c}{Full Target Data Methods} \\
\midrule
DA-Faster \cite{DAFRCNN} & Full & 41.9 \\
IRG \cite{IRGG} & Full & 43.2 \\
D\&Match \cite{DMatch} & Full & 43.9 \\
SWDA \cite{SWDA} & Full & 44.6 \\
DSS \cite{DSS} & Full & 44.5 \\
SC-DA \cite{SCDA} & Full & 45.1 \\
AFAN \cite{AFAN} & Full & 45.5 \\
MTOR \cite{MTOR} & Full & 46.6 \\
GPA \cite{GIPA} & Full & 47.6 \\
ViSGA \cite{ViSGA} & Full & 49.3 \\
MKT \cite{MKT} & Full & 50.2 \\
TDD \cite{TDD} & Full & 63.3 \\
\midrule
\multicolumn{3}{c}{Zero-Shot \& One-Shot Methods} \\
\midrule
Faster R-CNN \cite{frcnn} & Zero & 40.2\scriptsize$\pm$1.0\normalsize \\
NP \cite{NP} & Zero & 51.0 \\
OSFA (VGG16+SSD) \cite{OSFA} & One & 38.6 \\
OSSA (Ours) & One & \textbf{52.7}\scriptsize$\pm$0.5\normalsize \\
\bottomrule
\end{tabular}%
}
\vspace{0.3cm}
\caption{Comparison of different UDA methods on Sim10k to Cityscapes (sim2real) adaptation. Unless stated otherwise, all methods use a ResNet50+Faster R-CNN architecture.}
\label{tab:sim10k_comparison}
\end{table}

\FloatBarrier
\subsection{Visible-to-Thermal Adaptation}

In Table~\ref{tab:m3fd}, we observe the results of applying OSSA to the very different domain scenario of visible-to-thermal adaptation. We set a first benchmark for unsupervised one-shot adaptation in visible-to-thermal object detection. On this challenging task we show a substantial improvement over the baseline and the state-of-the-art domain generalization method NP. However, we note that most of the gap remains unbridged. Interestingly, this problem is observed in several papers on various datasets~\cite{meta-uda, feature, edgeguided}. This suggests that the scenario of visible-to-thermal adaptation remains an open problem. One possible explanation for the results is that certain elements that may be believed to be stylistic, such as the high intensity/brightness of car wheels, may be semantic and meaningful for an object detector which conflicts with our prior definition of style as semantically irrelevant. Hence, simply trying to establish an invariance or adapting the style at a global level, as done by OSSA, may be too inconsiderate of the nuanced differences between spectra.

\begin{table}[htbp]
\centering
\resizebox{\textwidth}{!}{%
\begin{tabular}{@{}lccccccc@{}}
\toprule
Method & Target Data & Person & Car & Truck & Bus & Motorcycle & mAP $\uparrow$ \\
\midrule
\rowcolor{lightgray} Oracle & Full & 66.8\scriptsize$\pm$0.4\normalsize & 80.7\scriptsize$\pm$0.3\normalsize & 69.7\scriptsize$\pm$1.1\normalsize & 82.0\scriptsize$\pm$1.5\normalsize & 54.5\scriptsize$\pm$1.4\normalsize & 70.8\scriptsize$\pm$0.6\normalsize \\
\midrule
\multicolumn{8}{c}{Zero-Shot \& One-Shot Methods} \\
\midrule
Faster R-CNN \cite{frcnn} & Zero & 20.4\scriptsize$\pm$0.9\normalsize & 38.4\scriptsize$\pm$1.3\normalsize & 12.5\scriptsize$\pm$2.3\normalsize & 30.9\scriptsize$\pm$3.0\normalsize & 7.6\scriptsize$\pm$1.4\normalsize & 22.0\scriptsize$\pm$0.9\normalsize \\
NP \cite{NP} & Zero & 28.4 & 48.9 & 16.4 & 40.0 & 9.1 & 28.6 \\
OSSA (Ours) & One & \textbf{30.7}\scriptsize$\pm$1.2\normalsize & \textbf{57.3}\scriptsize$\pm$0.9\normalsize & \textbf{24.7}\scriptsize$\pm$2.5\normalsize & \textbf{42.3}\scriptsize$\pm$3.2\normalsize & \textbf{14.6}\scriptsize$\pm$2.9\normalsize & \textbf{33.9}\scriptsize$\pm$1.1\normalsize \\
\bottomrule
\end{tabular}%
}
\vspace{0.3cm}
\caption{Comparison of different UDA methods on M\textsuperscript{3}FD visible-to-thermal adaptation. All methods use a ResNet50+Faster R-CNN architecture.}
\label{tab:m3fd}
\end{table}

\FloatBarrier

\section{Ablation Study}
\label{sec:ablation}
We conducted a series of ablation studies to evaluate the effectiveness of various design choices within our one-shot style adaptation method. First, we explored the impact of sourcing style prototypes from different domains. As shown in Table \ref{tab:prototypes}, selecting a random target style prototype proves substantially more effective than sampling from the source domain. This highlights the importance of sampling the style image from the intended target to achieve effective results. 

\begin{figure}[H]
  \centering
  \begin{tabular}{cc}
    \begin{subtable}[t]{0.48\textwidth}
      \centering
      \resizebox{\textwidth}{!}{%
        \begin{tabular}{@{}lcc@{}}
          \toprule
          & Cityscapes $\rightarrow$ Foggy & Sim10k $\rightarrow$ Cityscapes \\
          \midrule
          Target Prototypes & \textbf{40.1} & \textbf{53.1} \\
          Source Prototypes & 36.3 & 51.3 \\
          \bottomrule
        \end{tabular}
      }
      \caption{Effect of channel style prototype.}
      \label{tab:prototypes}
    \end{subtable} &
    \begin{subtable}[t]{0.48\textwidth}
      \centering
      \resizebox{\textwidth}{!}{%
        \begin{tabular}{@{}lccc@{}}
          \toprule
          & Cityscapes & Cityscapes $\rightarrow$ Foggy & Sim10k $\rightarrow$ Cityscapes \\
          \midrule
          Layer 1 & \textbf{50.2} & 39.6 & 51.2 \\
          Layer 2 & 49.2 & 37.9 & 51.4 \\
          Layer 3 & 49.1 & 35.3 & 48.1 \\
          Layer 1,2 & 49.4 & \textbf{40.1} & \textbf{53.1} \\
          Layer 1,2,3 & 49.2 & 39.3 & 53.0 \\
          \bottomrule
        \end{tabular}
      }
      \caption{Effect of method positions.}
      \label{tab:positions}
    \end{subtable} \\
    \begin{subtable}[t]{0.48\textwidth}
      \centering
      \resizebox{\textwidth}{!}{%
        \begin{tabular}{@{}lccc@{}}
          \toprule
          & Cityscapes & Cityscapes $\rightarrow$ Foggy & Sim10k $\rightarrow$ Cityscapes \\
          \midrule
          G(1, 0.0) & 47.8 & 35.3 & 48.6 \\
          G(1, 0.25) & \textbf{50.0} & 39.2 & 51.7 \\
          G(1, 0.50) & 48.8 & 38.9 & 53.0 \\
          G(1, 0.75) & 49.4 & \textbf{40.1} & \textbf{53.1} \\
          G(1, 1.00) & 47.6 & 39.9 & 52.5 \\
          \bottomrule
        \end{tabular}
      }
      \caption{Effect of noise intensity.}
      \label{tab:noise_types}
    \end{subtable} &
    \begin{subtable}[t]{0.48\textwidth}
      \centering
      \resizebox{\textwidth}{!}{%
        \begin{tabular}{@{}lccc@{}}
          \toprule
          & Cityscapes & Cityscapes $\rightarrow$ Foggy & Sim10k $\rightarrow$ Cityscapes \\
          \midrule
          25\% & 49.7 & 39.8 & 51.0 \\
          50\% & \textbf{49.4} & \textbf{40.1} & 52.9 \\
          75\% & 49.3 & 39.9 & \textbf{53.1} \\
          100\% & 44.7 & 38.4 & 50.7 \\
          \bottomrule
        \end{tabular}
      }
      \caption{Effect of augmentation probability.}
      \label{tab:probability}
    \end{subtable}
  \end{tabular}
  \caption{Analysis of OSSA's performance across different settings, including the influence of channel style prototypes, method application at various network layers, varying noise intensity levels, and augmentation probabilities.}
  \label{fig:ossa_analysis}
\end{figure}

We assess, which layers of the backbone were most beneficial for applying OSSA, the optimal intensity of noise, and the best augmentation probability. Our findings, detailed in Tables \ref{tab:positions}, \ref{tab:noise_types}, and \ref{tab:probability}, indicate that applying the method at layers 1 and 2, using a noise standard deviation of 0.75, and an augmentation probability of 50\% yields the best results.  However, it is possible to deviate from these exact parameters without much performance degradation depending on the specific task. 

\begin{figure}[htbp] 
\centering
\includegraphics[width=0.48\textwidth]{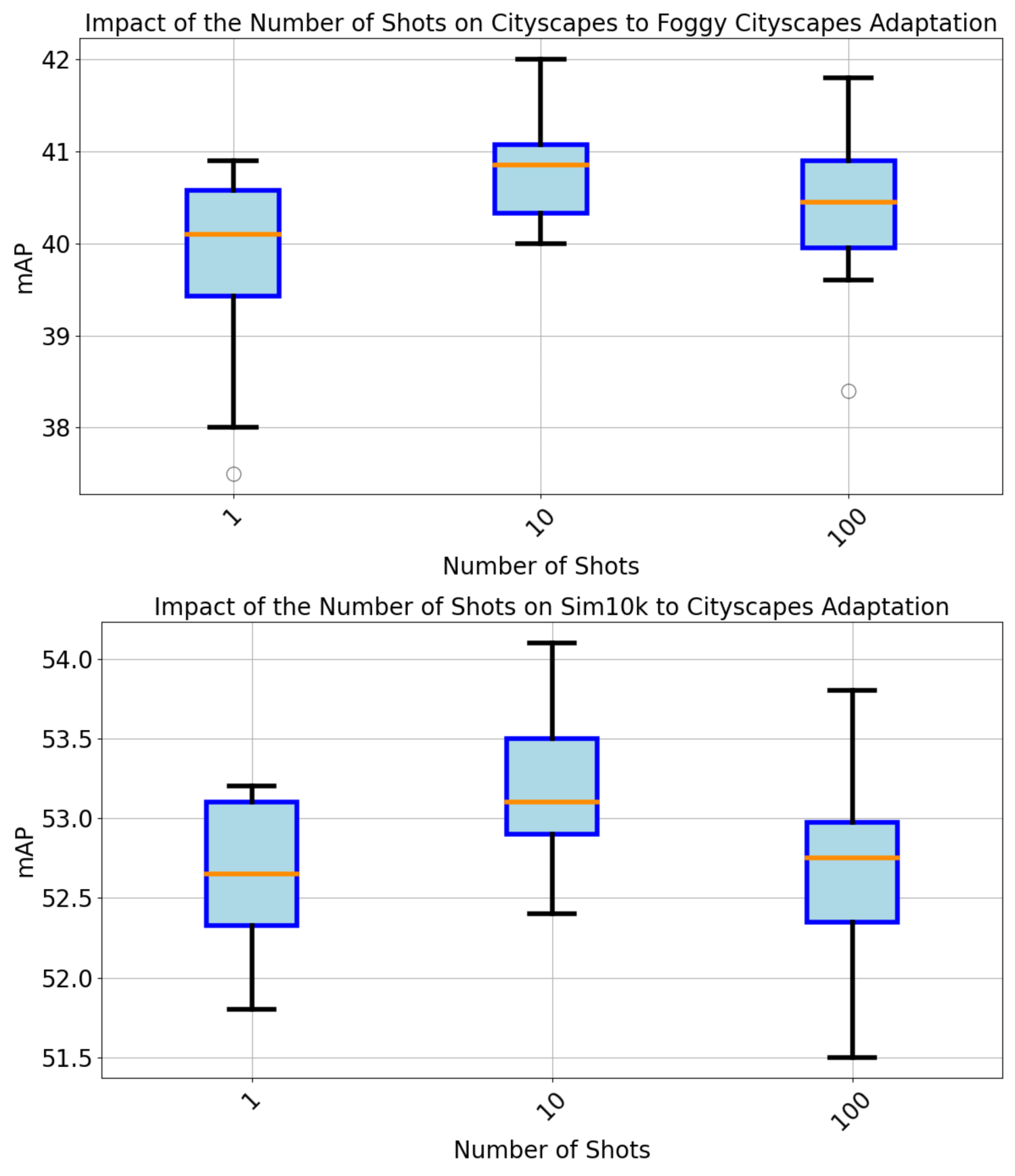} 
\caption{Boxplot of the impact of using multiple target images to approximate target channel style statistics. No significant increase in mean average precision is observed when utilizing more than one target image.}
\label{fig:boxes}
\end{figure}

Additionally, we investigated whether using a single target image to approximate the target style was sufficient, or if averaging across multiple images (10 or 100) would improve performance. The results, presented in Figure \ref{fig:boxes}, are mixed. In the adaptation from Cityscapes to Foggy Cityscapes, a slight improvement in mAP of approximately 0.5 is observed, although the large interquartile range suggests this is not significant. For the Sim10k to Cityscapes adaptation, increasing the sample size does not show a clear benefit. These outcomes suggest that a single target image is adequate for OSSA. If available, we suggest averaging over multiple samples to mitigate potential anomalies.

\section{Discussion and Conclusion}
\label{sec:conclusion}
This paper introduces a novel one-shot, unsupervised domain adaptation technique for object detection that effectively bridges the style gap between domains. Our method leverages style statistics at the feature level, enabling a rapid and effective adaptation process that significantly reduces the need for extensive target domain knowledge. Our approach has been tested on challenging benchmarks such as Cityscapes to Foggy Cityscapes and Sim10k to Cityscapes, in which it not only surpassed existing one-shot methods but also narrowed the gap to, or surpassed, strong baselines, which access thousands of target images when performing adaptation. 
The substantial performance gains achieved by our method, particularly notable in scenarios like the adaptation from Cityscapes to Foggy Cityscapes, where it approached 80\%  of the oracle's performance with just one image, highlight its ability to utilize minimal data to close the style gap. By demonstrating that competitive performance can be achieved with significantly reduced data, our research advocates for a shift toward more data-efficient practices in unsupervised adaptation.
Another key advantage of OSSA is its relative simplicity, contributing to its interpretability. This is a valuable trait in deep learning research, as it allows us to better understand the inner workings of the adaptation process and identify areas for improvement. Additionally, OSSA seamlessly integrates into existing frameworks by operating within the backbone network, requiring minimal architectural modifications. This ease of integration makes OSSA a practical candidate solution for real-world applications.
The versatility of our technique is further exemplified by its successful application across different domains. While some methods are specifically designed for certain types of style gaps, such as those created by different weather conditions~\cite{CLIP, Li_2023_WACV}, we show that OSSA is a more versatile approach. We demonstrated this by showing success in regular benchmarks such as weather and sim2real adaptation, as well as more nuanced fields such as visible-to-thermal adaptation. 

Looking ahead, we note that OSSA is specifically designed to bridge the style gap between domains. However, our observations reveal that the gap to the oracle remains significant in cases like Sim10k to Cityscapes or the M3FD visible-to-thermal benchmark. This highlights the need for future methods to address not only the style gap but also the content or semantic gaps between datasets, especially in extreme scenarios such as visible-to-thermal adaptation. Moreover, OSSA's potential application in Source-Free Domain Adaptation (SFDA) methods, such as those that rely on virtual domain generation \cite{li2024comprehensive}, presents an intriguing direction for future research. Additionally, future research should explore OSSA's effectiveness across various CNN backbones, ranging from established architectures like VGG16, known for its compatibility with AdaIN, to more modern architectures~\cite{10028728}, which could enhance data-efficient learning. Finally, investigating OSSA's integration with other image-level augmentations or batch normalization within complex systems could provide valuable insights for further performance improvements.

{
    \small
    \bibliographystyle{splncs04}
    \bibliography{main}
}
\end{document}